\documentclass{article}

\usepackage[main, final]{neurips_2026}
\usepackage{graphicx}
\usepackage{subcaption}
\usepackage{amsmath}
\usepackage{amssymb}
\usepackage{mathtools}
\usepackage{amsthm}
\usepackage{wrapfig}

\usepackage[utf8]{inputenc}
\usepackage[T1]{fontenc}
\usepackage{hyperref}
\usepackage{url}
\usepackage{booktabs}
\usepackage{amsfonts}
\usepackage{nicefrac}
\usepackage{microtype}
\usepackage{xcolor}
\usepackage{caption}

\title{$\text{A}^3$-VLA: Automatic Perception-Guided \\ Attention Alignment for Robot Manipulation}

\newcommand{\samethanks}{\footnotemark[\value{footnote}]}

\author{%
\textbf{Shengzhe Zhang}\textsuperscript{1}
\quad
\textbf{Qi Zhang}\textsuperscript{2}%
\thanks{Corresponding authors.}
\quad
\textbf{Dazhong Shen}\textsuperscript{3}
\\
\textbf{Chao Wang \textsuperscript{4,5}}\samethanks
\quad
\textbf{Shaopeng Zhai \textsuperscript{2}}
\quad
\textbf{Hui Xiong \textsuperscript{6,7}}\samethanks
\\[1ex]
{\normalfont\small
\textsuperscript{1}School of Computer Science and Technology
, University of Science and Technology of China}
\\
{\normalfont\small
\textsuperscript{2}Shanghai AI Laboratory}
\\
{\normalfont\small
\textsuperscript{3}College of Computer Science and Technology, Nanjing University of Aeronautics and Astronautics}
\\
{\normalfont\small
\textsuperscript{4}School of Artificial Intelligence and Data Science,
University of Science and Technology of China}
\\
{\normalfont\small
\textsuperscript{5}State Key Laboratory of Cognitive Intelligence,
University of Science and Technology of China}
\\
{\normalfont\small
\textsuperscript{6}Thrust of Artificial Intelligence,
The Hong Kong University of Science and Technology (Guangzhou)}
\\
{\normalfont\small
\textsuperscript{7}Department of Computer Science and Engineering,
The Hong Kong University of Science and Technology}
\\[1ex]
\texttt{owen\_zsz@mail.ustc.edu.cn},
\texttt{zhangqi.fqz@gmail.com},
\texttt{dazh.shen@gmail.com},
\\
\texttt{wangchaoai@ustc.edu.cn},
\texttt{zsp1197@163.com},
\texttt{xionghui@ust.hk}
}

\begin{document}

\maketitle

\begin{abstract}
Vision-Language-Action Models (VLAs) enable robots to perform manipulation tasks by leveraging Vision-Language Models (VLMs) to process visual observations and language instructions.
However, the action prediction objective itself lacks fine-grained task-specific perceptual supervision, making it prone to relying on task-irrelevant visual shortcuts, leading attention away from target objects and key interaction areas, thereby limiting its operational performance and generalization ability.
To address this, we propose Automatic Perception-Guided Attention Alignment for Robot Manipulation ($\text{A}^3$-VLA), a framework that derives task-specific visual guidance from the manipulation process itself. Specifically, $\text{A}^3$-VLA first generates heuristic candidate regions for target objects and the robot from interaction-induced visual changes, and refines them with a visual segmentation model offline. These regions serve as explicit perceptual guidance through two complementary objectives: an attention allocation loss that encourages sufficient focus on task-critical regions when language or robot state tokens query the visual input, and a contrastive learning loss that aligns the representations of candidate target regions with language instructions while separating them from task-irrelevant regions. Extensive experiments on simulated and real-world manipulation tasks demonstrate that $\text{A}^3$-VLA consistently improves multiple VLA backbones, with an average success rate gain of 6.5\% - 25.2\% across benchmarks.
\end{abstract}

\section{Introduction}
\label{sec:intro}
Embodied manipulation aims to enable robots to interact with the environment through learned policies to complete manipulation tasks \cite{diffusion, Aloha}.
A core challenge is enabling robots to understand both what to manipulate (from language instructions) and how to manipulate (from visual observations).
Recently, Vision-Language Models (VLMs) have achieved remarkable success in aligning visual and linguistic representations, demonstrating strong capabilities in tasks such as visual question answering and cross-modal reasoning \cite{vlm, paligemma, XVLM, CoGen}.
Building upon these advances, Vision-Language-Action Models (VLAs) \cite{RT2, Octo, pi0} extend VLMs to robot control by directly generating action sequences conditioned on visual observations and language instructions \cite{MemoryVLA, UniVLA, ReconVLA, E0}.

\begin{wrapfigure}{R}{0.5\textwidth}
\centering
\includegraphics[width=\linewidth]{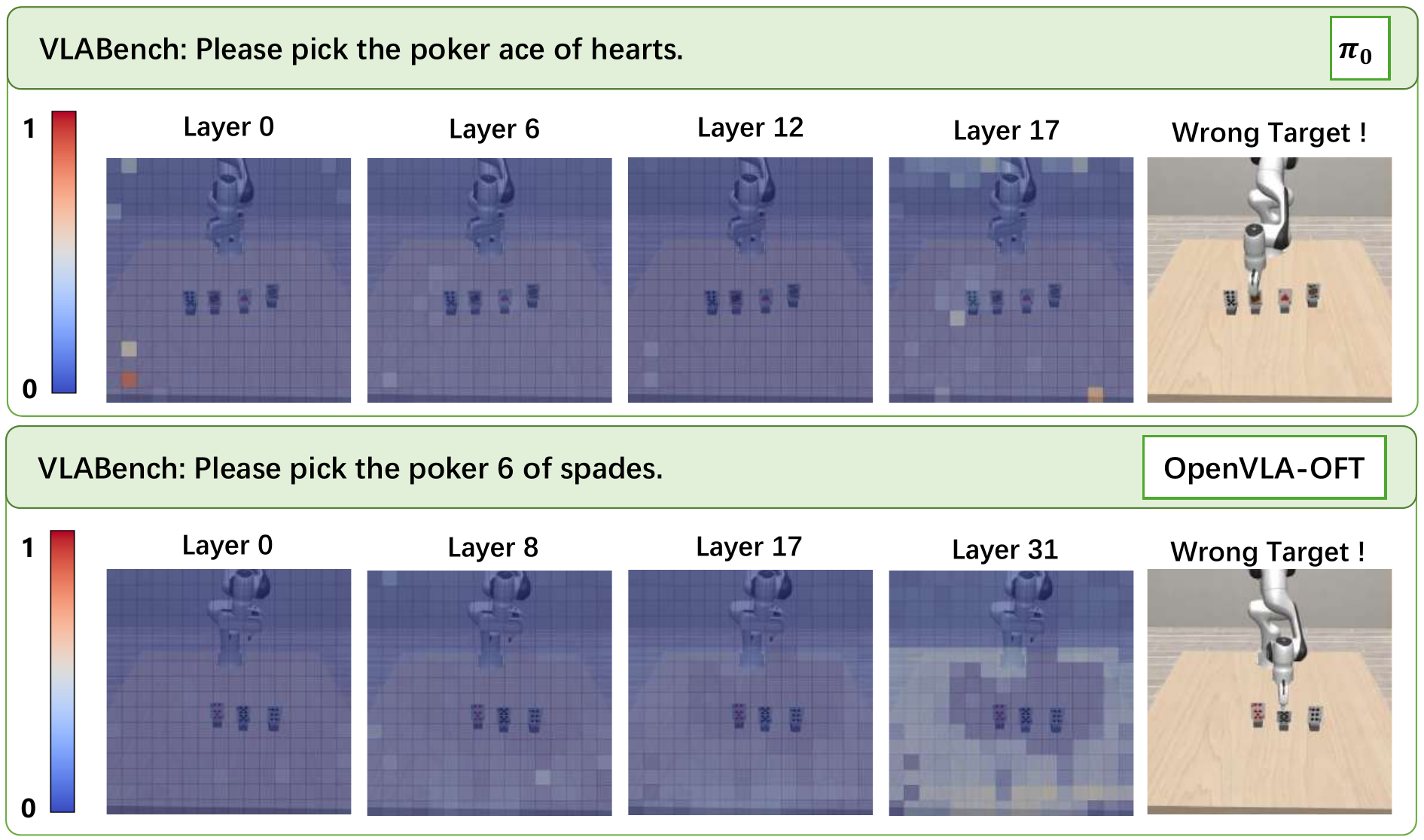}
\caption{
A failure case of the attention maps of language instruction querying visual observations in $\pi_0$ and OpenVLA-OFT on a Select Poker task from VLABench, visualized across layers from shallow (left) to deep (right).
}
\label{pics/intro-example}
\end{wrapfigure}

Early works on VLAs primarily focus on accurate action generation, pioneered by tokenization-based \cite{Octo, OpenVLA} and flow-matching-based \cite{pi0, Pi0-FAST} paradigms.
Despite this progress in action learning, the visual perception capability of these models, including their ability to attend to regions specified by language instructions, remains fragile. As shown in Figure~\ref{pics/intro-example}, these cases reveal a key failure mode that is often overlooked in VLA research: the model fails because the VLM backbone’s attention mechanism fails to focus on the task-relevant regions specified by the language instruction, rather than being solely attributable to action generation errors. Thus, the VLA learns to “memorize” specific action trajectories from the training data, rather than connecting the instructions with the observation of the target objects. To break this pattern, VLAs need explicit signals about where to look, raising a natural question: can we obtain such task-specific guidance from the manipulation process itself?

We argue that the above limitation reveals two deeper issues with how VLAs currently handle visual perception. First, the fine-tuning process provides no explicit signal indicating which image regions are critical for the given task. The flow-matching or autoregressive objective only supervises action accuracy \cite{HM4SR, VLA0, TGCL4SR}; whether the model's visual attention aligns with task-relevant regions is never directly evaluated or encouraged. As a result, the VLM backbone's pretrained ability to discriminatively attend to semantically meaningful regions can gradually drift during fine-tuning \cite{FACE, CEDAR, VLAC}. Second, prior works attempt to enhance perception in different ways, such as CoT-VLA \cite{CoTVLA} with visual chain-of-thought reasoning, $\pi_{0.5}$ \cite{Pi0.5} and Gemini Robotics \cite{Gemini} with open-world knowledge. The guidance they provide remains either indirect (reasoning over the entire scene) or task-agnostic (general knowledge from heterogeneous data).
These motivate a different perspective: the manipulation process itself encodes rich information about what matters visually. The moment of physical interaction, the trajectory of the robot end-effector, and the target object being acted upon all carry signals about which image regions deserve focused attention. If we obtain these signals and utilize them as explicit perceptual guidance, the task-oriented visual perception capability may be preserved during fine-tuning, rather than being overwritten by the action learning objective.

To address these challenges, we propose \textbf{A}utomatic Perception-Guided \textbf{A}ttention \textbf{A}lignment for Robot Manipulation ($\text{A}^3$-VLA), a framework that provides explicit visual perception guidance during VLA fine-tuning. The key insight of $\text{A}^3$-VLA is that interaction moments in the manipulation process naturally reveal the locations of task-critical regions. Based on this, $\text{A}^3$-VLA consists of two stages. First, a heuristic region proposal and temporal refinement stage extracts noisy candidate regions from interaction-induced visual changes and refines them with SAM2 into temporally consistent pseudo regions for target objects and the robot. Second, a perception-guided fine-tuning stage incorporates these pseudo regions via two complementary objectives: (i) an attention allocation loss that encourages the model to focus sufficient attention on the heuristic regions of target objects or the robot when language or robot state tokens query the visual input respectively, and (ii) a contrastive learning loss that aligns the representations of target object regions with language instructions while pushing away those of task-irrelevant regions. Together, these objectives help the VLA maintain task-focused visual attention throughout fine-tuning, rather than allowing it to drift toward task-irrelevant regions.
Overall, our contributions are as follows:
\begin{itemize}
\item We propose $\text{A}^3$-VLA, a framework that derives task-specific visual guidance from the manipulation process itself, eliminating manual annotation.
\item We design two auxiliary training objectives, attention allocation and contrastive representation alignment, that explicitly steer VLA visual focus toward task-relevant regions.
\item Extensive experiments on both simulation and real-world settings demonstrate that $\text{A}^3$-VLA consistently improves multiple VLA backbones across diverse manipulation tasks.
\end{itemize}

\section{Related Work}
In this section, we provide previous research relevant to our work, including Vision-Language-Action Models (VLAs) and multimodal perception.

\subsection{Vision-Language-Action Models (VLAs)}
Foundational works such as RT-1 \cite{RT1} and RT-2 \cite{RT2} pioneered action tokenization for end-to-end policy learning, followed by open-source VLAs including Octo \cite{Octo} and OpenVLA \cite{OpenVLA}, and diffusion/flow-matching approaches like Gr00t series \cite{GR00TN1} and $\pi_0$ \cite{pi0}. Recent extensions have explored efficient fine-tuning recipes \cite{XR_1, SmolVLA}, and lightweight adaptation modules \cite{VLA-Adapter} to enhance the efficiency of VLAs.
Across these methods, primary VLAs mainly supervise action learning yet overlook the significance of multimodal perception during fine-tuning \cite{ReconVLA}, which makes language following and task understanding remain challenging for VLAs \cite{Blind, SafeVLA}.

\subsection{Multimodal Perception for VLA}
VLMs have demonstrated strong multimodal perception capabilities in tasks including VQA \cite{InternVL, ChatVLA}, reasoning \cite{CoTVLA, VLM4VLA}, and cross-modal generation \cite{Qwen-Image, Llava}, etc.
Some works on VLAs attempt to explore and utilize the multimodal perception capability of VLMs to enrich the scene and task understanding of embodied models. For instance, RoboCLIP \cite{RoboCLIP} and Embodied CLIP \cite{EmbCLIP} ground language in robotic affordances with contrastive learning on the context of egocentric videos. FPC-VLA \cite{FPC-VLA} utilizes a fine-tuned supervisor to check for failure based on multimodal observations. Recently, $\pi_{0.5}$ \cite{Pi0.5} pretrains VLAs with heterogeneous data sources to transfer open-world multimodal knowledge. Gemini Robotics \cite{Gemini} harnesses Gemini's capacity for physical reasoning to advance spatial comprehension. Unlike prior methods that rely on general multimodal knowledge or indirect scene-level supervision, $\text{A}^3$-VLA automatically derives task-specific interaction cues from manipulation trajectories and refines them with a visual segmentation
model into soft pseudo regions for region-level perception guidance during fine-tuning.

\section{Methodology}

\begin{figure*}[t]
\centering
\includegraphics[width=\linewidth]{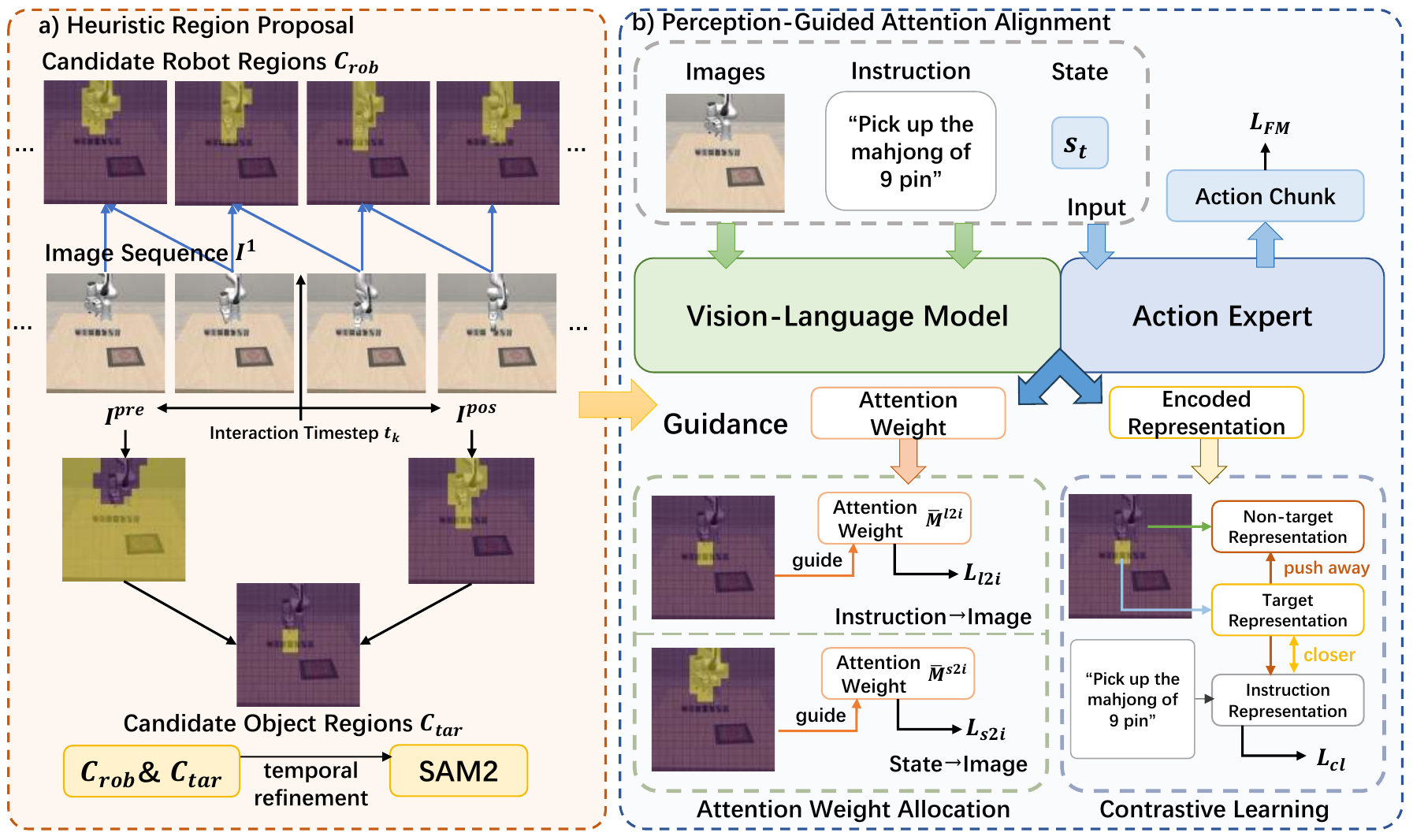}
\caption{Overview of the $\text{A}^3$-VLA framework.
}
\label{model}
\end{figure*}

In this section, we introduce the framework of Automatic Perception-Guided Attention Alignment for Robot Manipulation ($\text{A}^3$-VLA), which automatically derives pseudo-region guidance from the manipulation process and delivers explicit guidance for multimodal perception. Its framework is illustrated in Figure~\ref{model}. The key components of A3-VLA are: 1) \textbf{heuristic region proposal}, which extracts noisy candidate regions for target objects and the robot from demonstration trajectories and refines them with SAM2 \cite{SAM2}; 2) \textbf{attention weight allocation}, which encourages sufficient attention mass on pseudo regions; 3) \textbf{contrastive learning}, which aligns candidate target-region representations with language instructions; and 4) \textbf{joint VLA training} with the main action objective and auxiliary perceptual objectives.

\subsection{Preliminary}
The VLA flow model \cite{pi0} is a powerful and widely accepted paradigm for action generation, adopted as the backbone of our $\text{A}^3$-VLA. Formally, the observation at the timestep $t$ is denoted as $o_t=[I_1^t,...,I_n^t, l^t, s^t]$, where $I_i^t$ is the $i$-th view of images, $l^t$ is the language instruction, and $s^t$ is the current robot state. The main goal for flow matching-based VLAs is to learn the future action chunk $A_t=\{a_t, ..., a_{t+H}\}$, where $a_t$ and $H$ represent the action and the action horizon respectively. The corresponding flow matching loss $L_{FM}$ is formulated as follows:
\begin{equation}
    \scalebox{0.95}{$\displaystyle
    L_{FM}(\theta)=\mathbb{E}_{p(\mathbf{A}_{t}|\mathbf{o}_{t}),q(\mathbf{A}_{z}^{t}|\mathbf{A}_{t})}\left[\left\|\mathbf{f}_{\theta}(\mathbf{A}_{z}^{t},\mathbf{o}_{t})-\mathbf{u}(\mathbf{A}_{z}^{t}|\mathbf{A}_{t})\right\|^{2}\right],
    $}
\end{equation}
where $\theta$ is the parameter of the model, $\mathbf{A}_{z}^{t}$ is the noised action whose noise distribution is $q(\mathbf{A}_{z}^{t}|\mathbf{A}_{t}) = \mathcal{N}(z A_t, (1-z)I)$,
$\mathbf{f}_{\theta}(\mathbf{A}_{z}^{t},\mathbf{o}_{t})$ denotes the output of the VLA, and $\mathbf{u}(\mathbf{A}_{z}^{t}|\mathbf{A}_{t})$ is the denoising vector field in flow matching.

\subsection{Heuristic Region Proposal}
\label{la}
To provide explicit perceptual guidance during VLA fine-tuning, we first propose candidate regions that are likely to be task-relevant for the given manipulation task. Our key insight is that useful heuristic cues can be naturally extracted from the manipulation process itself. In many manipulation tasks, the robot physically interacts with task-relevant objects during the trajectory. These interaction moments, which are detectable from the robot states or sensors, provide direct signals for localizing both the target objects and the robot in the visual observation. In the following, we describe how interaction cues are used to generate candidate target and robot regions.

Formally, consider a trajectory with a sequence of images $I=\{I^1,..., I^T\}$ and corresponding robot state sequence $S=\{s^1, ..., s^T\}$, where $T$ is the trajectory length. The interaction timestep $t_k$ denotes a coarse moment when the robot starts interacting with the task-relevant object. In practice, such moments can be obtained from robot or task cues, e.g., gripper states, contact signals, or predefined state ranges. Once $t_k$ is determined, we partition image patches into two key region sets: the target object $C_{tar}$ and the robot $C_{rob}$. The two region types are identified through different procedures, reflecting their distinct roles in the manipulation process. For target object regions $C_{tar}$, to localize the target object, we leverage the fact that it remains in place before the robot makes contact but undergoes visual changes once interaction begins. We define a window of $w$ frames, and for each patch at spatial position $(i,j)$, we compute the mean absolute difference between consecutive frames within the pre-interaction window $[t_k-2w,t_k-w]$ and the post-interaction window $[t_k,t_k+w]$ respectively:
\begin{equation}
d^{pre}_{(i,j)} = \frac{1}{w}\sum_{t=t_k-2w}^{t_k-w}||I^{t+1}_{(i,j)} - I^{t}_{(i,j)}||_2,
\end{equation}
\begin{equation}
d^{post}_{(i,j)} = \frac{1}{w}\sum_{t=t_k}^{t_k+w}||I^{t+1}_{(i,j)} - I^{t}_{(i,j)}||_2,
\end{equation}
where $I^{t}_{(i,j)}$ denotes the RGB value of patch $(i,j)$ at the $t$-th frame. This heuristic relies on the observation that the target object is often visually stable before physical interaction and changes after being manipulated. Then, patches with low pre-interaction variation but high post-interaction variation, i.e., $d^{pre}_{(i,j)}<\tau$ and $d^{post}_{(i,j)}>\tau$, are treated as candidate regions for the target object $C_{tar}$. These patches are visually stable before interaction but change once the robot begins interacting, matching the expected behavior of an object being manipulated. For the robot regions $C_{rob}$, unlike target objects, it is present and active throughout the entire trajectory. It is therefore identified by finding patches with substantial variation across nearby observations:
\begin{equation}
d^{t}_{(i,j)} = ||I^{t+w}_{(i,j)} - I^{t}_{(i,j)}||_2.
\end{equation}
Then, patches with high variation, i.e., $d^{t}_{(i,j)}>\tau$, are treated as candidate regions $C_{rob}$. Here $\tau$ is a variation threshold. We use $\tau=10$ across all experiments and find performance is insensitive to this choice.
The above description assumes a single interaction per trajectory for clarity. For tasks involving multiple interactions, the trajectory is partitioned at each interaction timestep, regions are obtained for each segment using the same method, and the final label sets are taken as the union across all segments.

Notably, a key practical advantage of this heuristic region proposal approach is that it requires no manual labeling, since it operates solely on the demonstration trajectory. These heuristic proposals are not intended to be accurate object masks. Instead, they serve as high-recall pseudo regions that provide coarse task-relevant visual cues. The asymmetric criterion, i.e., low variation before interaction and high variation after interaction, helps suppress patches that are consistently dynamic or unrelated to the contact event. For multi-interaction tasks, we apply the same procedure within each interaction segment before taking the union of the resulting regions, which reduces the ambiguity caused by temporally separated object changes.

Since the heuristic proposals may still be noisy or incomplete, we further propagate them across frames using SAM2 to obtain temporally consistent region tracks. This refinement serves two purposes: (i) smoothing out frame-to-frame inconsistencies of the heuristic cues, and (ii) propagating the initial proposals to frames both before and after the interaction moment, providing region coverage for the entire trajectory without requiring the heuristic to work perfectly on every frame. Therefore, the labeling process only requires coarse interaction-induced seeds, rather than precise frame-level annotations. More details are provided in Appendix~\ref{Sam2}.

\subsection{Guidance with Attention Weight Allocation}
\label{ga}

The heuristic regions $C_{tar}$ and $C_{rob}$ obtained in Section~\ref{la} provide cues about where the VLA should look. While these heuristic regions are not guaranteed to be perfectly accurate, they provide a useful attention prior that we exploit through soft constraints rather than hard labels. Here, we adopt a soft guidance strategy: rather than hard-constraining the attention distribution, we impose only a lower bound on the total attention allocated to the pseudo regions, encouraging VLAs to preserve focus on task-critical areas while retaining the freedom for action learning and remaining visual knowledge, without disrupting the visual understanding capabilities that make the VLM backbone valuable in the first place. Here, we use the attention map from the last layer of VLMs and average over all attention heads.

\textbf{Language-to-image attention alignment.} We first build the bridge between the semantics of the language instruction and the visual understanding with the guidance of $C_{tar}$. Concretely, let $M^{l2i}$ denote the normalized attention weight matrix from the $N_L$ language tokens to the image patch tokens. We first obtain the average attention weight matrix:
\begin{equation}
\overline{M}^{l2i}_{(i,j)} = \frac{1}{N_l} \sum_{m=1}^{N_l} M^{l2i}_{m,(i,j)}.
\end{equation}
Then, we calculate how much weight is assigned to the region of the target item patches, and align the semantics of language and vision by encouraging enough attention weight for target items, which can be formulated as follows:
\begin{equation}
    L_{l2i} = [ max(0, \eta_{1} - \sum_{(i,j)\in C_{tar}} \overline{M}^{l2i}_{(i,j)}) ]^{2} ,
\end{equation}
where $\eta_{1}$ denotes the minimum threshold for target item attention weight assignments, encouraging VLAs to maintain focus on the instruction-specified object without over-constraining action learning.

\textbf{State-to-image attention alignment.} While $L_{l2i}$ connects language to the target objects, another alignment is needed between the robot state and the robot visual embodiment. Accurate action generation depends on the model's ability to relate its state to the spatial configuration of the robot in the image. Using the candidate robot label $C_{rob}$ and the normalized attention weight matrix $M^{s2i}$ from the state token to the image patch tokens, we apply an analogous constraint:
\begin{equation}
    L_{s2i} = [ max(0, \eta_{2} - \sum_{(i,j)\in C_{rob}} M^{s2i}_{(i,j)}) ]^{2} ,
\end{equation}
where $\eta_{2}$ denotes the minimum attention mass encouraged on candidate robot regions. Together, $L_{l2i}$ and $L_{s2i}$ provide explicit, region-specific guidance that directly counteracts the tendency of VLA attention to drift toward task-irrelevant regions during fine-tuning. By softly anchoring the model's attention to the regions that physically matter for the task, these objectives help the VLA retain focused visual perception throughout fine-tuning, directly addressing the attention drift problem identified in our motivating examples.

\subsection{Guidance with Contrastive Learning}
The guidance in Section~\ref{ga} provides supervision at the level of attention weights, while explicitly shaping the semantic relationship between language instructions and the visual features extracted from those regions is also required. To strengthen this relationship, we turn to the representation space. Our key insight is that the regions $C_{tar}$ from Section 3.2 naturally partition each pre-interaction frame into two semantically distinct sets: patches inside the refined candidate target region are treated as pseudo-positive visual evidence, while patches outside the target regions are treated as context negatives. The former should be semantically close to the language instruction, while the latter should not. Based on this, we design a contrastive learning objective that takes the language instruction representation as an anchor, pulls the visual features of $C_{tar}$ toward it, and pushes the features of non-target regions away. This provides patch-level supervision that explicitly aligns the model's visual encoding of the target object with its linguistic understanding of the task.
Formally, let $P^{lan}$ denote the encoded representations of the $N_l$ language instruction tokens, and $P^{img}$ be the encoded representations of the image patch tokens from the VLA. We adopt the mean pooling strategy to obtain the representations of the semantic anchor, positive visual parts and negative visual parts:
\begin{equation}
\overline{P}_{lan} = \frac{1}{N_l} \sum_{m=1}^{N_l} P^{lan}_m,
\overline{P}^{+}_{img} = \frac{1}{|C_{tar}|}\sum_{(i,j) \in C_{tar}} P^{img}_{(i,j)},
\overline{P}^{-}_{img} = \frac{1}{N_P-|C_{tar}|}\sum_{(i,j) \notin C_{tar}} P^{img}_{(i,j)},
\end{equation}
where $N_P$ is a fixed number of image patches consistent with the image processing method of the VLM backbone. Then the formulation of contrastive learning is as follows:
\begin{equation}
    \scalebox{0.93}{$\displaystyle
    L_{cl} = -\log \frac{\exp(sim(\overline{P}_{img}^{+}, \overline{P}_{lan}))}{\exp(sim(\overline{P}_{img}^{+}, \overline{P}_{lan})) + \exp(sim(\overline{P}_{img}^{-}, \overline{P}_{lan}))},
    $}
\end{equation}
where $sim(,)$ denotes the cosine similarity function. We apply $L_{cl}$ only before interaction, where the target object is typically visible and less occluded by the robot, making language-target alignment cleaner.

\subsection{VLA jointly training}
With the procedures above, we have three training objectives for VLAs: $L_{FM}$ for action generation learning, $L_{l2i}$  and $L_{s2i}$ for explicit attention weight allocation guidance, and $L_{cl}$ for target-oriented representation enhancement. Together, these two forms of supervision address the perception gap from complementary angles: attention allocation shapes the model's visual focus on task-critical regions, while contrastive alignment ensures the visual features of those regions are semantically aligned with the language instruction.
Finally, $\text{A}^3$-VLA is jointly optimized with these tasks with the following formulation:
\begin{equation}
    L = L_{FM} + \lambda_1 (L_{l2i} + L_{s2i}) + \lambda_2 L_{cl} ,
\end{equation}
where $\lambda_1$ and $\lambda_2$ are the weights of loss functions. It's notable that our framework is model-agnostic and can be seamlessly adapted to diverse VLAs. Notably, during inference, A3-VLA uses only the fine-tuned VLA policy, thus with no additional computational cost.

\section{Experiment}
In this section, we introduce details of the experiments on the evaluation of $\text{A}^3$-VLA. We first compare our method with state-of-the-art VLA baselines on both simulation and real-world environments. Then, the ablation study and training efficiency analysis are conducted to display the impact of the designed components. Last, we present the case study to demonstrate how perception guidance improves the performance of VLAs.

\subsection{Experiment Setup}
\textbf{Simulation Benchmarks.} For simulation, we evaluate our methods on VLABench \cite{VLABench} and Robotwin 2.0 \cite{Robotwin, Robotwin2} benchmarks. VLABench provides a large scale of tasks to test the understanding of VLAs for semantic instructions and scene analysis,
while Robotwin is a widely adopted benchmark with a large variety of scenes and objects for bimanual robotic simulation.
For VLABench, we train and test models on five representative task suites: Chemistry Tube, Fruit, Poker, Toy and Mahjong, each containing 500 demonstrations. We test the VLAs over 50 episodes on each task suite in VLABench.
For Robotwin, we compare the VLAs on four tasks: Beat Block Hammer, Click Bell, Move Playingcard Away, and Place Burger \& Fries. For each task, we utilize Robotwin pipeline to generate 100 demonstrations for training, and test VLAs over 100 episodes.
In each environment, tasks are all jointly trained.

\noindent \textbf{Real-World Environment.} In the real-world experiments, we set up scenes of manipulation tasks in the kitchen with
a Songling robotic arm.
The tasks include Open Cooker, Pick \& Place Bowl, Pick 2 Vegetables, Select Bottles, and Desktop Sweep, each containing 80 demonstrations.
Evaluation for each task is conducted 20 times, and we further record the task progress score, which is set as 1 if the task is completed, 0.5 if the VLA fails halfway, and 0 for inconsistency with instructions.
In the real-world experiments, tasks are all trained together.

\noindent \textbf{Implementation Details.}
We implement $\text{A}^3$-VLA on $\pi_0$, OpenVLA-OFT, and $\pi_{0.5}$, adopting each backbone's original training settings.
For hyperparameters, the window size $w$ is the same as the frame rate of videos. The variance threshold $\tau$ is set as 10, while the attention weight thresholds $\eta_1$ and $\eta_2$ are both 0.25. The loss weights $\lambda_1$ and $\lambda_2$ are set as 0.1 and 0.01 respectively.
All the training and experiments are conducted on a server with 8 NVIDIA H200 GPUs.

\noindent \textbf{Baselines.}
We compare $\text{A}^3$-VLA with various state-of-the-art methods. Octo \cite{Octo} and OpenVLA \cite{OpenVLA} are pioneering works for open-source VLAs with OpenX pretraining. RDT \cite{RDT-1B} is a VLA with unified action space for transferable physical knowledge. $\pi_0$ \cite{pi0} and $\pi_0$-FAST \cite{Pi0-FAST} are flow-based VLAs that complete manipulation tasks by action chunk generation. We also evaluate the top-performing OpenVLA-OFT \cite{OpenVLA-OFT} and $\pi_{0.5}$ \cite{Pi0.5}. OpenVLA-OFT is an improved version of OpenVLA which improves action learning by the optimized fine-tuning recipe. $\pi_{0.5}$ is an enhanced version of $\pi_0$ which incorporates open world knowledge from heterogeneous sources of data.

\begin{table*}[]
\renewcommand{\arraystretch}{1.0}
\centering\small
\caption{Success Rate of VLAs on VLABench. The scores of the best baselines are underlined, while the scores of $\text{A}^3$-VLA are bold.}
\resizebox{0.85\linewidth}{!}{
\begin{tabular}{lcccccc}
\toprule
Method  & Chemistry & Fruit & Poker & Toy  & Mahjong & Avg.                  \\
\midrule
Octo\cite{Octo}    & 0.02           & 0.00   & 0.00   & 0.02 & 0.04    & 0.02                 \\
OpenVLA \cite{OpenVLA} & 0.08           & 0.06  & 0.04  & 0.10  & 0.06    & 0.07                 \\
RDT \cite{RDT-1B}   & 0.12           & 0.16  & 0.08  & 0.10  & 0.06    & 0.10                 \\
$\pi_0$-FAST \cite{Pi0-FAST} & 0.20              & 0.34     & 0.18     & 0.22    & 0.10       & 0.21                     \\
\midrule
$\pi_0$ \cite{pi0}     & \underline{0.52}           & \underline{0.60}   & 0.54  & \underline{0.40}  & 0.30     & \underline{0.47}                  \\
$\text{A}^3$-VLA ($\pi_0$)    & \textbf{0.82}              & \textbf{0.66}     & \textbf{0.82}     & \textbf{0.46}    & \textbf{0.42}       & \textbf{0.64} \\
\midrule
OpenVLA-OFT \cite{OpenVLA-OFT} & 0.32              & 0.18     & 0.14     & 0.12    & 0.04       & 0.16                     \\
$\text{A}^3$-VLA (OpenVLA-OFT) & \textbf{0.62} & \textbf{0.50} & \textbf{0.22} & \textbf{0.36} & \textbf{0.30} & \textbf{0.40} \\
\midrule
$\pi_{0.5}$ \cite{Pi0.5} & 0.30              & 0.42     & \underline{0.64}     & 0.32    & \underline{0.40}       & 0.41                     \\
$\text{A}^3$-VLA ($\pi_{0.5}$) & \textbf{0.74} & \textbf{0.50} & \textbf{1.00} & \textbf{0.42} & \textbf{0.68} & \textbf{0.67} \\
\bottomrule
\end{tabular}
}
\label{vlabench-main}
\end{table*}

\begin{table}
\centering
\captionof{table}{Success Rate of VLAs on RoboTwin 2.0. The scores of the best baselines are underlined, while the scores of $\text{A}^3$-VLA are bold.}
\resizebox{0.75\linewidth}{!}{
\begin{tabular}{lccccc}
\toprule
Method                         & \begin{tabular}[c]{@{}c@{}}Block\\ Hammer\end{tabular} & \begin{tabular}[c]{@{}c@{}}Click\\ Bell\end{tabular} & \begin{tabular}[c]{@{}c@{}}Playing\\ Card\end{tabular} & \begin{tabular}[c]{@{}c@{}}Burger\\ \& Fries\end{tabular} & Avg.          \\ \midrule
OpenVLA                        & 0.00                                                    & 0.09                                                  & 0.14                                                    & 0.00                                                       & 0.06           \\
RDT                            & 0.17                                                    & 0.44                                                  & 0.30                                                    & 0.08                                                       & 0.23           \\ \midrule
$\pi_0$                        & 0.29                                                   & \underline{0.56}                                                 & 0.70                                                   & 0.12                                                      & 0.42          \\
$\text{A}^3$-VLA ($\pi_0$)     & \textbf{0.36}                                          & \textbf{0.65}                                        & \textbf{0.79}                                          & \textbf{0.26}                                             & \textbf{0.52} \\ \midrule
OpenVLA-OFT                    & 0.06                                                   & 0.26                                                 & 0.39                                                   & 0.10                                                      & 0.20          \\
$\text{A}^3$-VLA (OpenVLA-OFT) & \textbf{0.08}                                          & \textbf{0.37}                                        & \textbf{0.46}                                          & \textbf{0.18}                                             & \textbf{0.27} \\ \midrule
$\pi_{0.5}$                    & \underline{0.56}                                                   & 0.44                                                 & \underline{0.81}                                                   & \underline{0.52}                                                      & \underline{0.58}          \\
$\text{A}^3$-VLA ($\pi_{0.5}$) & \textbf{0.59}                                          & \textbf{0.51}                                        & \textbf{0.89}                                          & \textbf{0.60}                                             & \textbf{0.65} \\ \bottomrule
\end{tabular}
}
\label{robotwin-main}
\end{table}

\begin{table*}[]
\renewcommand{\arraystretch}{1.1}
\centering\small
\caption{Success Rate (SR) and Task Progress (TP) of VLAs on Real-world Settings. Bold scores represent the highest results among all the methods, while the second highest scores are underlined.}
\resizebox{1.0\linewidth}{!}{
\begin{tabular}{lcccccccccccc}
\toprule
            & \multicolumn{2}{c}{\begin{tabular}[c]{@{}c@{}}Open\\ Cooker\end{tabular}} & \multicolumn{2}{c}{\begin{tabular}[c]{@{}c@{}}Pick \& Place\\ Bowl\end{tabular}} & \multicolumn{2}{c}{\begin{tabular}[c]{@{}c@{}}Pick 2 \\ Vegetables\end{tabular}} & \multicolumn{2}{c}{\begin{tabular}[c]{@{}c@{}}Select\\ Bottles\end{tabular}} & \multicolumn{2}{c}{\begin{tabular}[c]{@{}c@{}}Desktop\\ Sweep\end{tabular}} & \multicolumn{2}{c}{Avg.} \\ \midrule
Method      & SR                                  & TP                                  & SR                                      & TP                                     & SR                                  & TP                                 & SR                                   & TP                                   & SR                                   & TP                                   & SR           & TP           \\ \midrule
$\pi_0$ \cite{pi0}        &  \underline{0.85}                                &  \underline{0.93}                                &  \underline{0.75}                                    &  \underline{0.88}                                   &  0.40                               &  0.58                             &  0.25                                &  0.30                               &  0.50                                & 0.65                                 & 0.55         & 0.67         \\
$\pi_{0.5}$ \cite{Pi0.5}      &  \textbf{0.90}                                   &  \textbf{0.95}                    &             0.65   &                                      0.83   &                                                                           \underline{0.45}  &                                  \underline{0.60}   &                                    \underline{0.40}   &                             \underline{0.40}   &     \underline{0.70}     &                                   \underline{0.85}   &                                     \underline{0.62}  &    \underline{0.73}                        \\ \midrule
$\text{A}^3$-VLA ($\pi_0$)        &  \textbf{0.90}                & \textbf{0.95}           &        \textbf{0.80}  &                               \textbf{0.90}                                                                                                                      &                                \textbf{0.55}    &                                \textbf{0.78}     &                                   \textbf{0.75}   &                                  \textbf{0.80}   &       \textbf{0.80}  &                           \textbf{0.90}    &           \textbf{0.76}  &  \textbf{0.87}        \\ \bottomrule
\end{tabular}
}\label{realworld-main}
\end{table*}

\begin{table*}[t]
\centering

\begin{minipage}[t]{0.48\textwidth}
\centering
\setlength{\tabcolsep}{1.5mm}
\renewcommand{\arraystretch}{1.12}
\captionof{table}{The ablation Study of $\text{A}^3$-VLA on VLABench. C/F/P/T/M denote five tasks of VLABench respectively.}
\resizebox{\linewidth}{!}{
\begin{tabular}{lcccccc}
\toprule
Variations & C    & F    & P    & T    & M    & Avg. \\ \midrule
$\text{A}^3$-VLA ($\pi_0$)      & \textbf{0.82} & \textbf{0.66} & \textbf{0.82} & \textbf{0.46} & \textbf{0.42} & \textbf{0.64}    \\
w/o $L_{l2i}$      & 0.62 & 0.60 & 0.60 & 0.42 & 0.34 & 0.52    \\
w/o $L_{s2i}$      & 0.76 & 0.62 & 0.78 & 0.46 & 0.38 & 0.60    \\
w/o $L_{cl}$      & 0.78 & 0.66 & 0.80  & 0.40 & 0.40 & 0.61    \\
$\pi_0$         & 0.52 & 0.60 & 0.54 & 0.40 & 0.30 & 0.47    \\ \bottomrule
\end{tabular}
}
\label{ablation}

\end{minipage}
\hfill
\begin{minipage}[t]{0.48\textwidth}
\setlength{\tabcolsep}{1.5mm}
\centering
\setlength{\tabcolsep}{1mm}
\captionof{table}{Evaluation under texture (TP) and Gaussian noise (NP) perturbations on VLABench.}
\resizebox{\linewidth}{!}{
\begin{tabular}{lcccccc}
\toprule
Method & C    & F    & P    & T    & M    & Avg. \\ \midrule
$\pi_{0.5}$      & \textbf{0.30} & \textbf{0.42} & \textbf{0.64} & \textbf{0.32} & \textbf{0.40} & \textbf{0.41}    \\
+ TP      & 0.18 & 0.04 & 0.20 & 0.26 & 0.20 & 0.18    \\
+ NP      & 0.26 & 0.24 & 0.60 & 0.26 & 0.30 & 0.33    \\ \midrule
$\text{A}^3$-VLA ($\pi_{0.5}$)      & \textbf{0.74} & \textbf{0.50} & \textbf{1.00} & \textbf{0.42} & \textbf{0.68} & \textbf{0.67}    \\
+ TP      & 0.60 & 0.10 & 0.66 & 0.30 & 0.46 & 0.42    \\
+ NP      & 0.62 & 0.32 & 0.94 & 0.36 & 0.60 & 0.57    \\ \bottomrule
\end{tabular}
}
\label{perturb}

\end{minipage}
\end{table*}

\subsection{Comparison Results}
\textbf{VLABench Results.}
The results of experiments on VLABench are shown in Table~\ref {vlabench-main}.
In general, the performance of $\text{A}^3$-VLA successfully surpasses all baselines. Compared to the backbones $\pi_{0}$, OpenVLA-OFT, and $\pi_{0.5}$, $\text{A}^3$-VLA gains average 16.4\%, 24.0\%, and 25.2\% success rate improvements respectively, showing the consistent advancement of our method across different VLA frameworks. Specifically, $\text{A}^3$-VLA ($\pi_{0}$) improves higher in selecting Chemistry tube, Poker, and Mahjong compared to $\pi_0$, achieving 30\%, 28\%, and 12\% of performance gains respectively.
The improvement is particularly pronounced on tasks with visually similar distractors (e.g., Poker, Mahjong), demonstrating the effectiveness of our perception guidance for fine-grained object discrimination.

\noindent \textbf{Robotwin Results.}
 We evaluate our approach and other baselines on Robotwin 2.0, and report the quantitative results in Table~\ref{robotwin-main}. As shown in Table 2, $\text{A}^3$-VLA also improves all backbones on RoboTwin 2.0, with average gains of 9.8\%, 7.0\%, and 6.5\% percentage points. The largest gains appear on Place Burger \& Fries, suggesting that perception guidance is helpful for longer-horizon sequential manipulation.

\noindent \textbf{Real-world Results.}
We make real-world experiments and compare the performance of our approach with $\pi_0$ and $\pi_{0.5}$ on various kitchen-themed manipulation tasks. The results are shown in Table~\ref {realworld-main}, which demonstrates the advancement of $\text{A}^3$-VLA to complete different real-world tasks. Overall, our method outperforms the strongest baseline $\pi_{0.5}$ with 14\% average success rates and 14\% average task progress score. Among these tasks, $\text{A}^3$-VLA gains the most significant improvement on Select Bottles, which is a task demanding VLAs to pick up the specified bottle. This indicates that our design is helpful for VLAs' instruction understanding in the real-world background. By contrast, without perception-guided attention alignment, $\pi_0$ and $\pi_{0.5}$ have difficulty selecting the correct bottles. However in general, $\pi_{0.5}$ performs better than $\pi_0$ in various tasks.

\subsection{Ablation Study}
To verify the effectiveness of our designs, we conduct an ablation study on the key components of $\text{A}^3$-VLA. The results are displayed in Table~\ref {ablation}, which demonstrates that our designs are consistently effective for robot manipulation. Among these training objectives, the improvement of $L_{l2i}$ is the most significant, dropping 12\% average success rate when removing it, since it enhances semantic understanding and visual analysis more directly by aligning the attention map of language querying images with target item patches. $L_{cl}$ and $L_{s2i}$ bring moderate advancements, dropping 3\% and 4\% average success rate respectively when blocking them, as they benefit representation learning and alignment between vision and robot states respectively. Overall, these auxiliary training tasks can jointly boost the performance and the capability of multimodal perception for VLAs.

\subsection{Perturbation Robustness}
For robustness, we test $\text{A}^3$-VLA towards two kinds of perturbations: 1) Texture perturbations (TP) test VLAs by changing the environments into unseen textures. 2) Noise perturbations (NP) test VLAs by adding Gaussian noise to the input images. We set the standard deviation of the Gaussian noise as $10/255$. The results are shown in Table ~\ref {perturb}. $\text{A}^3$-VLA remains stronger than $\pi_{0.5}$ under both perturbations, suggesting that the learned perception guidance improves robustness without using any external module at test time. Examples of this study are in Appendix \ref{pe-a}.

\begin{figure}
\centering
\includegraphics[width=0.9\linewidth]{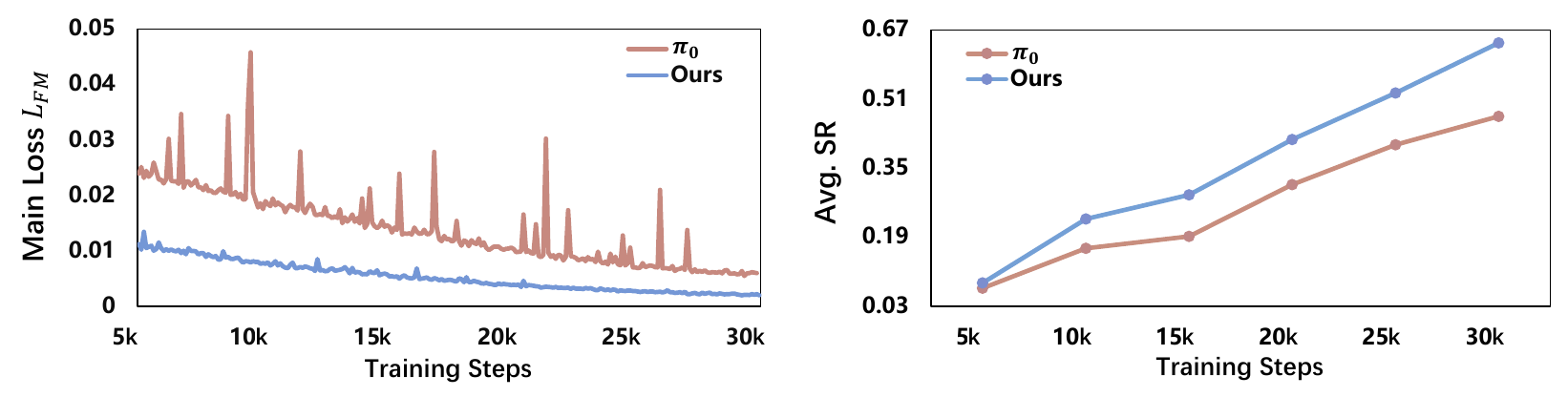}
\caption{The comparison of $L_{FM}$ and progressive SR of $\pi_0$ and $\text{A}^3$-VLA ($\pi_0$) on VLABench.}
\label{te}
\end{figure}

\subsection{Training Efficiency}
To validate the effect of $\text{A}^3$-VLA on the action learning progress, we illustrate the value of $L_{FM}$ during training and evaluate their performance every 5k training steps for $\pi_0$ and $\text{A}^3$-VLA, which are shown in Figure~\ref {te}. It shows a clear performance gap between $\text{A}^3$-VLA and the backbone $\pi_0$ under identical training settings. Specifically, $\text{A}^3$-VLA not only learns action policies more rapidly but also exhibits a smoother and more stable convergence curve, indicating a more robust learning process. This trend suggests that the auxiliary perceptual objectives may facilitate action learning by providing structured visual grounding. $L_{l2i}$ encourages language-target association, $L_{s2i}$ links robot states with visual embodiment, and $L_{cl}$ strengthens language-target alignment, reducing the burden of finding task-relevant cues solely from the action loss.

\subsection{Case Study}
\begin{figure}[t]
\centering
\includegraphics[width=\linewidth]{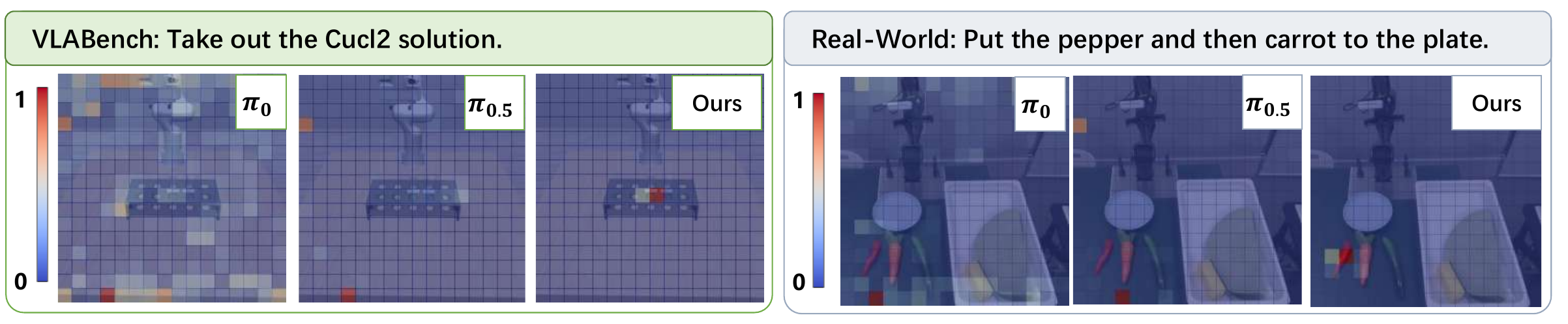}
\caption{The visualization of the attention map $\overline{M}^{l2i}$ for language querying the image when testing $\pi_0$, $\pi_{0.5}$ and $\text{A}^3$-VLA ($\pi_0$) on different tasks and benchmarks.}
\label{case}
\end{figure}

To achieve a deeper analysis of the effect of multimodal perception guidance, we further visualize the attention map $\overline{M}^{l2i}$ of $\text{A}^3$-VLA to investigate the alignment result for attention allocation guidance, and compare it with $\pi_0$ and $\pi_{0.5}$, illustrated in Figure ~\ref{case}. Through these cases, we can have the following observations. Firstly, for $\text{A}^3$-VLA, the regions with the highest attention weights largely correspond to the target objects in the language instructions. This demonstrates the effectiveness of semantic alignment between language and visual observations with explicit target object patch guidance. Compared with $\pi_0$ and $\pi_{0.5}$, $\text{A}^3$-VLA concentrates attention more consistently around instruction-specified targets across both simulated and real-world tasks, supporting the effect of perception-guided attention alignment.

\section{Conclusion}
In this paper, we proposed $\text{A}^3$-VLA, a multi-modal perception-enhanced VLA framework that improved task-oriented visual understanding for robot manipulation. Specifically, to incorporate manipulation-progress information into language-visual understanding, $\text{A}^3$-VLA automatically generated heuristic candidate regions for target objects and the robot from manipulation trajectories, which were then temporally refined into pseudo regions using SAM2. To guide VLAs to allocate sufficient attention to task-critical pseudo regions and learn higher-quality semantic representations, we proposed two alignment training objectives: one encouraged attention weights to concentrate on candidate target or robot regions when processing language commands and observations, while the other used contrastive learning to bring target-region representations closer to instructions. Extensive experiments on both simulated and real-world environments validated the effectiveness of the proposed method.

\section{Acknowledgments}
This work was supported in part by
New Generation Artificial Intelligence-National Science and Technology Major Project (Grant No. 2025ZD0122601), the National Natural Science Foundation of China (Grant No. 62506348), the Natural Science Foundation of Anhui Province (Grant No. 2508085QF211), and the Opening Foundation of State Key Laboratory of Cognitive Intelligence, iFLYTEK (COGOS-2025HE02),
the National Natural Science Foundation of China (Grant No.92370204), the National Key R\&D Program of China (Grant No.2023YFF0725001), Guangdong Provincial Key Laboratory of Frontier Basic Science for All‑domain Intelligence, the guangdong Basic and Applied Basic Research Foundation (Grant No.2023B1515120057), the Key-Area Special Project of Guangdong Provincial Ordinary Universities (2024ZDZX1007),
and the National Natural Science Foundation of China (Grant No. 62506185).

\section{Competing Interests}
The authors declare no competing interests.

\bibliographystyle{plain}
\bibliography{ref}

\appendix

\clearpage
\setcounter{page}{1}
\setcounter{section}{0}

\section{Appendix}

\subsection{Parameter Sensitivity}
\begin{figure*}[hb]
\centering
\includegraphics[width=\linewidth]{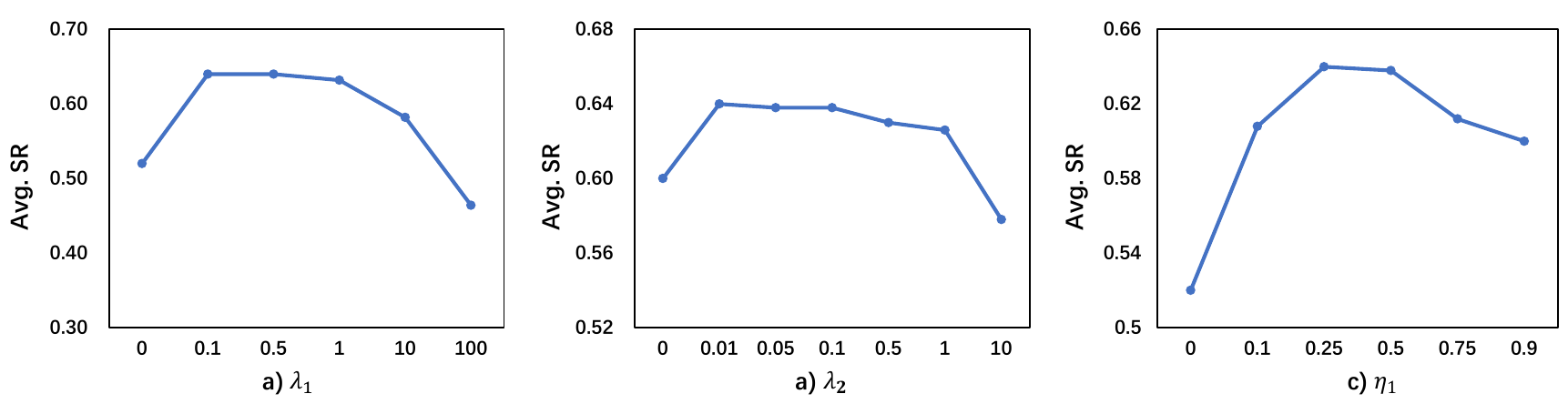}
\caption{Performance comparison on VLABench of $\text{A}^3$-VLA w.r.t different hyper-parameters.}
\label{ps}
\end{figure*}

In this part, we study the impact of important hyper-parameters of $\text{A}^3$-VLA, including the loss weight of the guidance with attention weight allocation $\lambda_1$, the loss weight of contrastive learning $\lambda_2$, and the threshold of attention weight allocation for language querying images $\eta_1$. When changing one hyper-parameter, we keep other hyper-parameters fixed to control variables. The results are illustrated in Figure~\ref{ps}. First, for $\lambda_1$, we can see that a lighter weight for the guidance with attention weight allocation is sufficient to improve the performance significantly. Whereas as the weight increases, the performance of $\text{A}^3$-VLA drops, probably because $\text{A}^3$-VLA gets too stuck to attention weight allocation and overlooks the action learning from $L_{FM}$. $\lambda_2$ shows the similar tendency as $\lambda_1$. Next, for $\eta_1$, the results demonstrate that the guidance with attention weight allocation is the most powerful for the VLA enhancement when $\eta_1$ has a moderate value. As it gets larger, the VLA may pay too much attention to the target regions while ignoring whole scene information, slightly weakening the improvement from the guidance with attention weight allocation.

\subsection{Examples for Perturbation Robustness}
\label{pe-a}
\begin{figure*}[h]
\centering
\includegraphics[width=\linewidth]{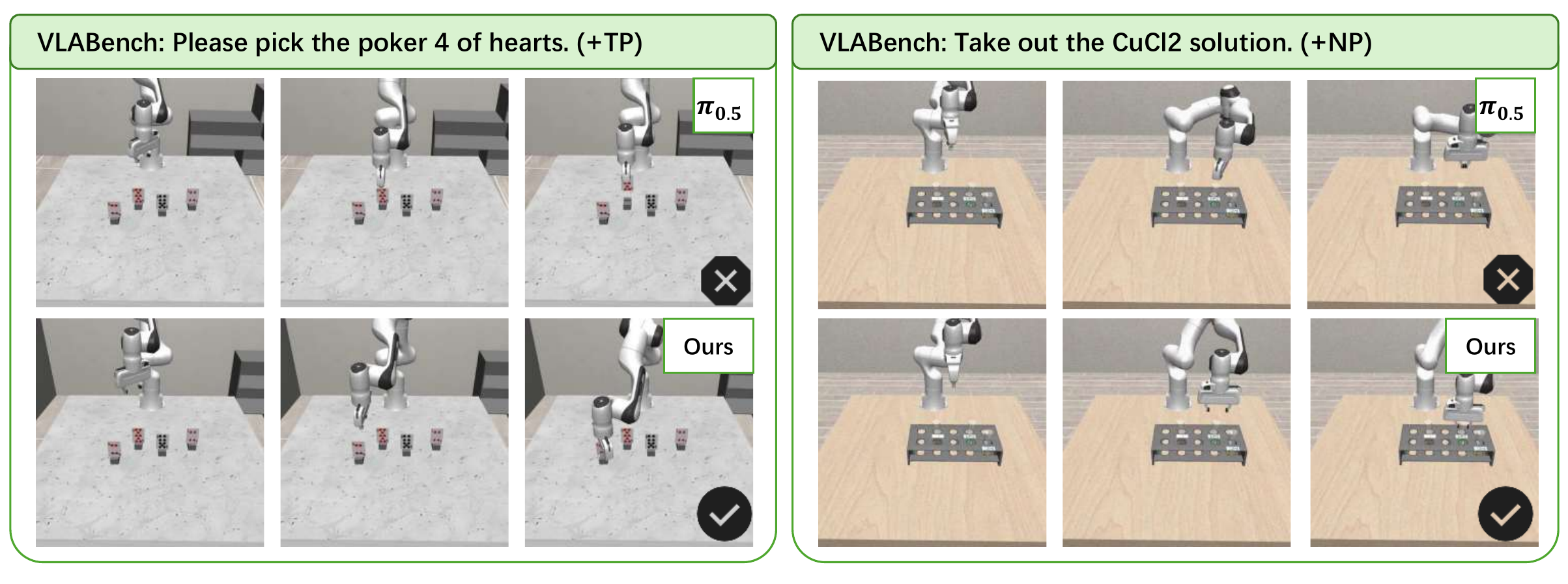}
\caption{The manipulation process of $\text{A}^3$-VLA ($\pi_{0.5}$) and $\pi_{0.5}$ under perturbations on VLABench.}
\label{pe-p}
\end{figure*}

We visualize examples for the manipulation process of $\text{A}^3$-VLA ($\pi_{0.5}$) and $\pi_{0.5}$ under texture and noise perturbations in Figure \ref{pe-p}. We can see that $\text{A}^3$-VLA ($\pi_{0.5}$) still performs better, showing our robustness to perturbations such as background texture changing.

\subsection{More Examples for Case Study}
\begin{figure*}
\centering
\includegraphics[width=0.9\linewidth]{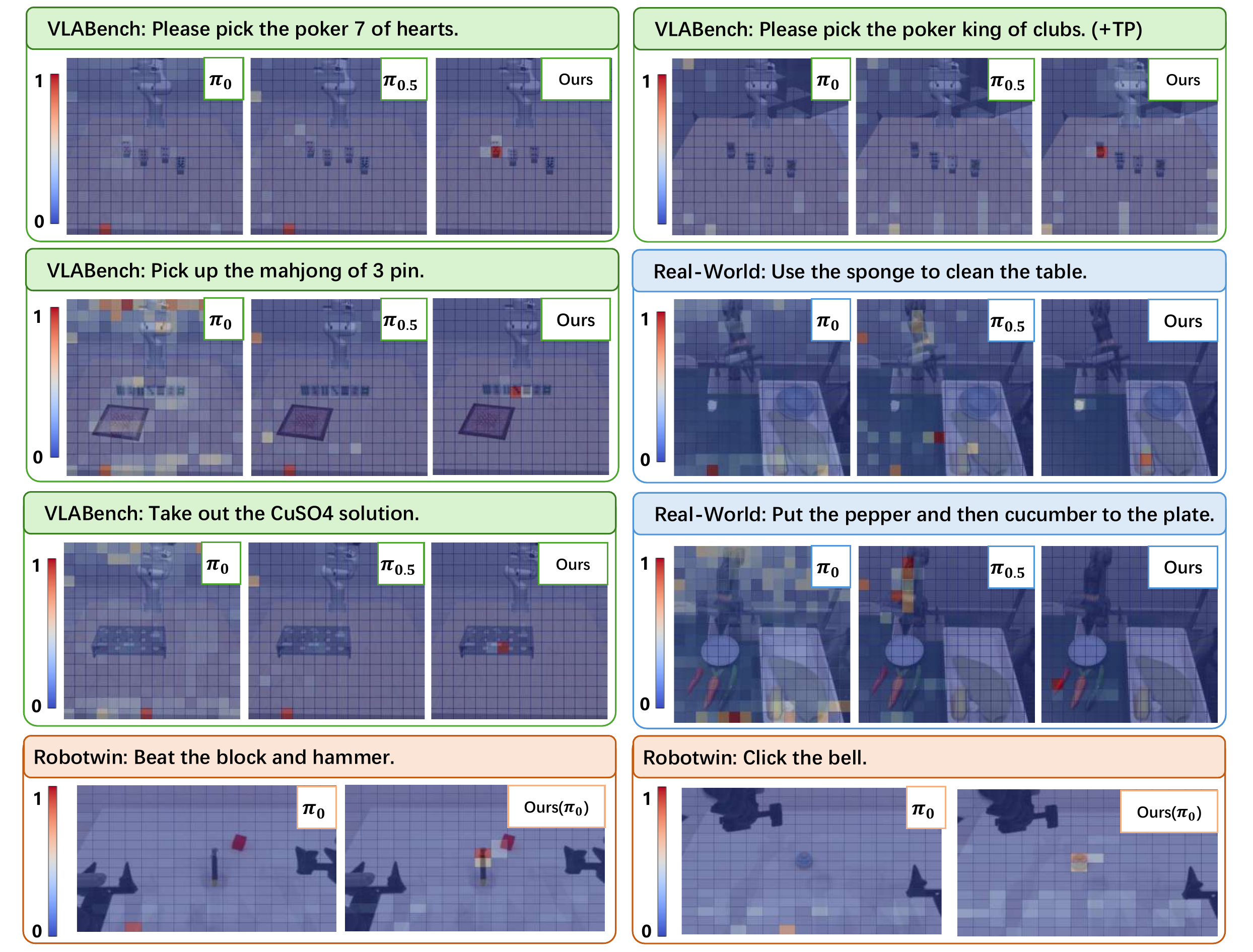}
\caption{The visualization of the attention map $\overline{M}^{l2i}$ for language querying the image when testing $\pi_0$, $\pi_{0.5}$ and $\text{A}^3$-VLA ($\pi_{0}$) on different tasks and benchmarks.}
\label{MCS}
\end{figure*}

We present more examples of the case study in Figure~\ref{MCS}. The conclusion is consistent: the regions with the highest attention weights of $\text{A}^3$-VLA largely correspond to the target objects specified by the language instructions. The attention allocation of $\pi_0$ is quite blurry, which is scattered throughout the whole image. Whereas $\pi_{0.5}$ sticks to specific positions of patches. In the last case, it's observed that the attention except for the pepper shifts from the carrot to the cucumber compared with the last case in Figure~\ref{case}, showing the semantic understanding of $\text{A}^3$-VLA. Further more, under the unseen texture perturbations (TP), $\text{A}^3$-VLA is still able to find out the regions of the target object, showing our robustness.

\subsection{More Examples for Heuristic Region Proposal}
\begin{figure*}
\centering
\includegraphics[width=0.9\linewidth]{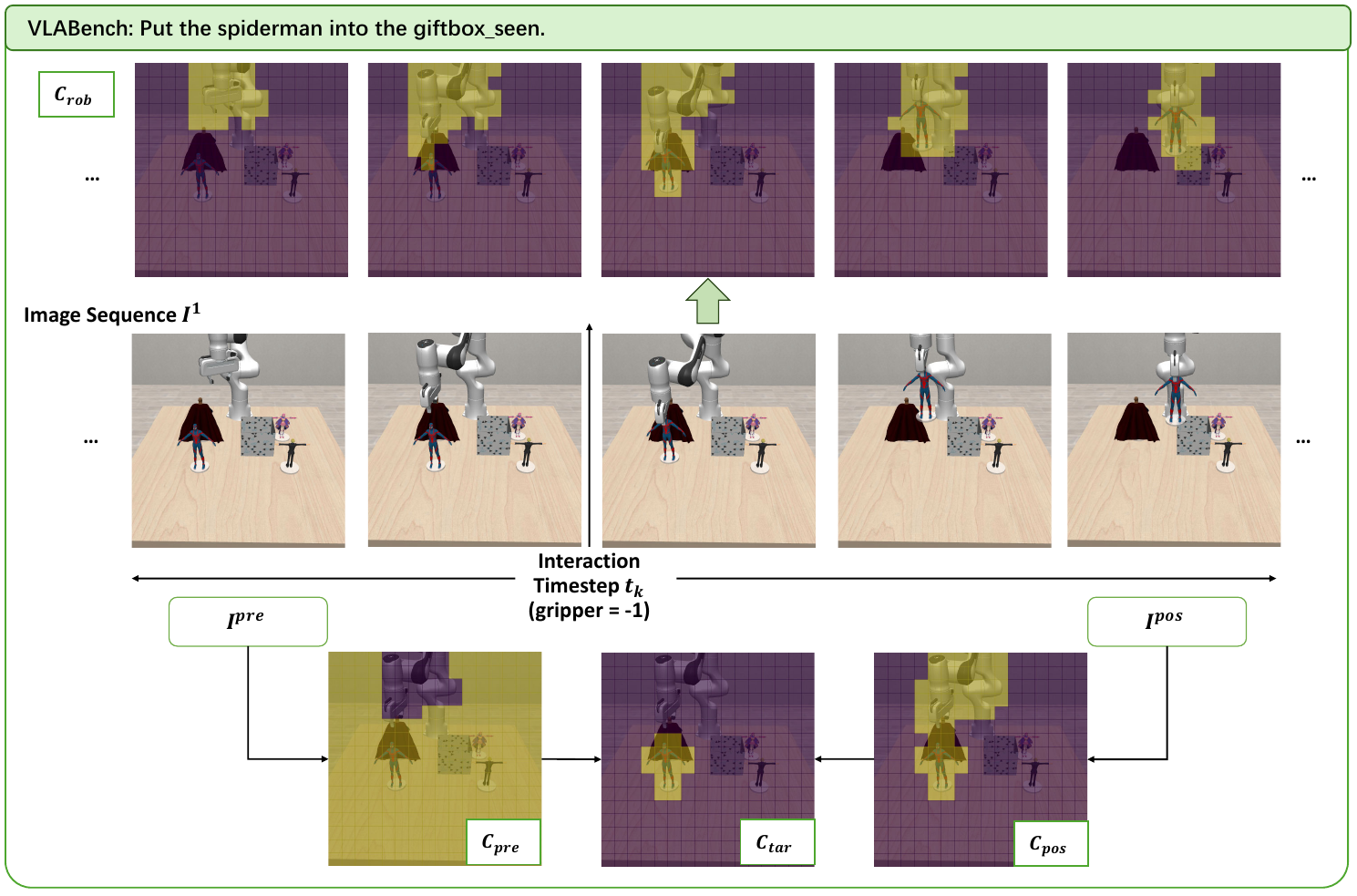}
\caption{An Example of heuristic region proposals on VLABench.}
\label{MIPL1}
\end{figure*}

\begin{figure*}
\centering
\includegraphics[width=0.9\linewidth]{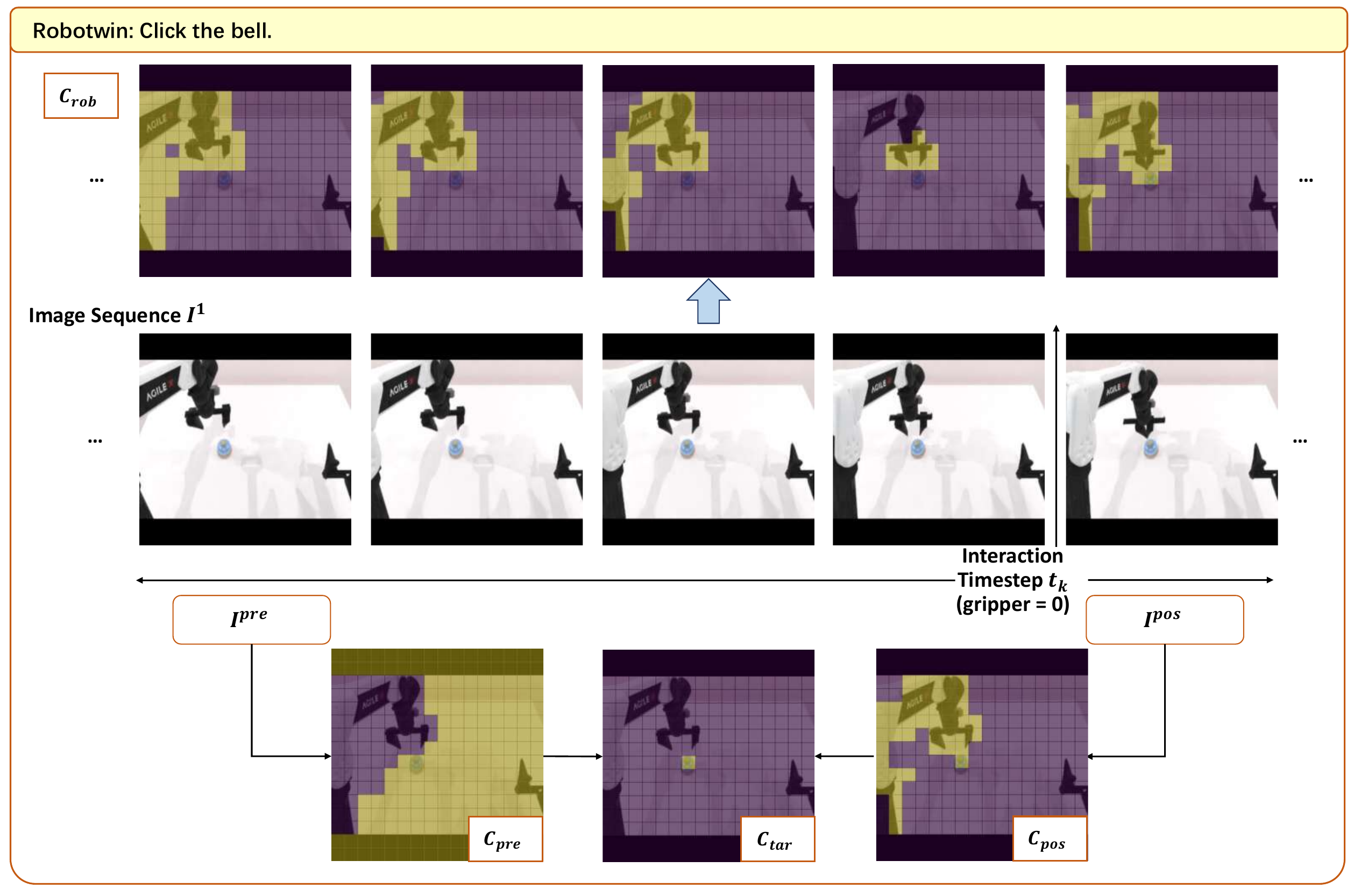}
\caption{An Example of heuristic region proposals on Robotwin.}
\label{MIPL2}
\end{figure*}

In this section, we show more examples of the results from the heuristic region process in Figure~\ref{MIPL1} and Figure~\ref{MIPL2}. The first task is to put the spiderman into the gift box, while the second task is to click the bell. Though in different benchmarks, our method can localize candidate regions of target items and robots across different situations and environments. This indicates that our method is useful under different situations and environments.

\subsection{More Details of Implementation}
The training batch sizes of $\pi_0$ and $\pi_{0.5}$ are set as 32 and 256 respectively, and they are both trained in 30k steps. For OpenVLA-OFT, we set the batch size and training steps as 8 per GPU and 50k, respectively.
For the choice of $\tau$, we find that $\tau=10$ works well across datasets, and slight changes to this value do not noticeably affect the results. Therefore, the heuristic region proposal is not sensitive to this threshold.
All baselines are fine-tuned with the dataset of the benchmarks.

\subsection{Details of SAM2-based Temporal Refinement}
\label{Sam2}
We use SAM2 as an off-the-shelf temporal refinement tool to improve the consistency of the heuristic region proposals. Given the candidate target and robot regions obtained from interaction-induced visual changes, we first convert the selected image patches at the interaction frame into binary masks. These binary masks are directly used as mask prompts for SAM2. Target and robot regions are prompted and refined separately using their corresponding heuristic masks.
Starting from the prompted interaction frame, SAM2 propagates the masks along the demonstration trajectory to produce temporally consistent pseudo masks for each frame. This refinement helps reduce isolated noisy patches and provides region estimates for frames where the original heuristic cue is weak or unavailable. The resulting masks are not treated as ground-truth annotations; they are used only as pseudo regions for soft perceptual supervision during VLA fine-tuning.
We further map each refined pixel-level mask to the VLA patch grid. Let $M^t_{tar}$ and $M^t_{rob}$ denote the refined target and robot masks at timestep $t$, and let $\Omega$ denote the set of image patches. A patch is assigned to the target or robot pseudo region if its overlap ratio with the corresponding mask exceeds a threshold $\rho$:
\begin{equation}
C^t_{tar}
=
\left\{
p \in \Omega
\mid
\frac{|p \cap M^t_{tar}|}{|p|}
>
\rho
\right\},
\end{equation}
\begin{equation}
C^t_{rob}
=
\left\{
p \in \Omega
\mid
\frac{|p \cap M^t_{rob}|}{|p|}
>
\rho
\right\}.
\end{equation}
The patch-level pseudo regions $C^t_{tar}$ and $C^t_{rob}$ are then used in the attention allocation losses and the contrastive objective. All pseudo masks are precomputed and cached before policy training. Therefore, SAM2 introduces additional offline preprocessing cost but no extra computation during policy inference. $\rho$ is empirically set as 0.25.

\subsection{Limitations}
$\text{A}^3$-VLA relies on interaction-induced visual changes to obtain heuristic candidate regions.
Although these regions are further refined by SAM2 and used only as soft supervision, the quality of the pseudo regions may degrade when the target object is fully occluded, visually indistinguishable from the background, too small to produce reliable patch-level changes, or moves together with distractors.
In such cases, the auxiliary perception losses may provide weaker or noisier guidance.

Another limitation is that our current implementation assumes that a coarse interaction moment can be obtained from lightweight robot states or task-state cues, such as gripper states, contact signals, or predefined state ranges.
While such cues are commonly available in manipulation demonstrations, more complex embodiments or tasks without clear interaction events may require more robust interaction discovery mechanisms.

$\text{A}^3$-VLA also introduces an offline preprocessing step using SAM2 to improve the temporal consistency of pseudo regions.
This does not add inference-time cost, since SAM2 is not used during policy execution, but it increases data preprocessing time and introduces a dependence on the quality of the off-the-shelf segmentation model.
Future work may explore uncertainty-aware weighting of pseudo regions, more robust interaction-guided region discovery, and fully end-to-end perception guidance without external refinement tools.

\subsection{Broader Impact}
Positive impacts include improving robot instruction following and reducing manual annotation for robot learning. Potential risks include failures in real-world deployment due to incorrect perception or attention alignment, which may cause unsafe robot actions. $\text{A}^3$-VLA should therefore be deployed with safety constraints, human supervision, and task-specific validation in physical environments.

\newpage
\section*{NeurIPS Paper Checklist}

\begin{enumerate}

\item {\bf Claims}
    \item[] Question: Do the main claims made in the abstract and introduction accurately reflect the paper's contributions and scope?
    \item[] Answer: \answerYes{}
    \item[] Justification: The abstract and introduction clearly summarize our contributions and scope. We focus on VLAs and propose a framework to protect the vision ability of their VLM backbone.

\item {\bf Limitations}
    \item[] Question: Does the paper discuss the limitations of the work performed by the authors?
    \item[] Answer: \answerYes{}
    \item[] Justification: We discuss the limitations in the Appendix. By explicitly acknowledging these limitations, we provide a balanced view of our work and suggest directions for future work.

\item {\bf Theory assumptions and proofs}
    \item[] Question: For each theoretical result, does the paper provide the full set of assumptions and a complete (and correct) proof?
    \item[] Answer: \answerNA{}
    \item[] Justification: Our method has no theoretical results.

    \item {\bf Experimental result reproducibility}
    \item[] Question: Does the paper fully disclose all the information needed to reproduce the main experimental results of the paper to the extent that it affects the main claims and/or conclusions of the paper (regardless of whether the code and data are provided or not)?
    \item[] Answer: \answerYes{}
    \item[] Justification: We provide the details of benchmarks, implementation, hyperparameter settings, and pseudo-region generation in the paper and Appendix.

\item {\bf Open access to data and code}
    \item[] Question: Does the paper provide open access to the data and code, with sufficient instructions to faithfully reproduce the main experimental results, as described in supplemental material?
    \item[] Answer: \answerYes{}
    \item[] Justification: We give the URL of our code. The datasets for VLABench and Robotwin are publicly available.

\item {\bf Experimental setting/details}
    \item[] Question: Does the paper specify all the training and test details (e.g., data splits, hyperparameters, how they were chosen, type of optimizer) necessary to understand the results?
    \item[] Answer: \answerYes{}
    \item[] Justification: We specify all the training and test details in the paper and the Appendix.

\item {\bf Experiment statistical significance}
    \item[] Question: Does the paper report error bars suitably and correctly defined or other appropriate information about the statistical significance of the experiments?
    \item[] Answer: \answerNo{}
    \item[] Justification: Each task is evaluated dozens of times, and we report the average success rate for consistency and reliability.

\item {\bf Experiments compute resources}
    \item[] Question: For each experiment, does the paper provide sufficient information on the computer resources (type of compute workers, memory, time of execution) needed to reproduce the experiments?
    \item[] Answer: \answerYes{}
    \item[] Justification:  The information on compute resources, including GPUs, is provided in the paper.

\item {\bf Code of ethics}
    \item[] Question: Does the research conducted in the paper conform, in every respect, with the NeurIPS Code of Ethics \url{https://neurips.cc/public/EthicsGuidelines}?
    \item[] Answer: \answerYes{}
    \item[] Justification: Our research conforms with NeurIPS Code of Ethics in every aspect.

\item {\bf Broader impacts}
    \item[] Question: Does the paper discuss both potential positive societal impacts and negative societal impacts of the work performed?
    \item[] Answer: \answerYes{}
    \item[] Justification: We discuss both potential positive societal impacts and negative societal impacts of our works in the Appendix.

\item {\bf Safeguards}
    \item[] Question: Does the paper describe safeguards that have been put in place for responsible release of data or models that have a high risk for misuse (e.g., pre-trained language models, image generators, or scraped datasets)?
    \item[] Answer: \answerNA{}
    \item[] Justification: In our experiments, we employ publicly available embodiment datasets that are widely used in the research community, together with established open-source VLMs that naturally comply with open-science principles. Because these datasets are non-sensitive and the underlying methods are fully transparent, we consider the risk of misuse to be negligible.

\item {\bf Licenses for existing assets}
    \item[] Question: Are the creators or original owners of assets (e.g., code, data, models), used in the paper, properly credited and are the license and terms of use explicitly mentioned and properly respected?
    \item[] Answer: \answerYes{}
    \item[] Justification: We adhere rigorously to these standards by guaranteeing that all assets used in our research—including code, datasets, and pre-trained models—are properly credited via references to their original authors.

\item {\bf New assets}
    \item[] Question: Are new assets introduced in the paper well documented and is the documentation provided alongside the assets?
    \item[] Answer: \answerYes{}
    \item[] Justification: We offer our code in an anonymous GitHub repository.

\item {\bf Crowdsourcing and research with human subjects}
    \item[] Question: For crowdsourcing experiments and research with human subjects, does the paper include the full text of instructions given to participants and screenshots, if applicable, as well as details about compensation (if any)?
    \item[] Answer: \answerNA{}
    \item[] Justification: Our research contains neither crowdsourcing experiments nor research with human subjects.

\item {\bf Institutional review board (IRB) approvals or equivalent for research with human subjects}
    \item[] Question: Does the paper describe potential risks incurred by study participants, whether such risks were disclosed to the subjects, and whether Institutional Review Board (IRB) approvals (or an equivalent approval/review based on the requirements of your country or institution) were obtained?
    \item[] Answer: \answerNA{}
    \item[] Justification: Our research contains neither crowdsourcing experiments nor research with human subjects.

\item {\bf Declaration of LLM usage}
    \item[] Question: Does the paper describe the usage of LLMs if it is an important, original, or non-standard component of the core methods in this research? Note that if the LLM is used only for writing, editing, or formatting purposes and does \emph{not} impact the core methodology, scientific rigor, or originality of the research, declaration is not required.
    \item[] Answer: \answerYes{}
    \item[] Justification: VLMs are utilized as a backbone for VLAs.

\end{enumerate}

\end{document}